\documentclass{INTERSPEECH2023}
\usepackage{color}

\interspeechcameraready

\title{CN-Celeb-AV: A Multi-Genre Audio-Visual Dataset for Person Recognition}
\name{Lantian Li$^2$, Xiaolou Li$^2$, Haoyu Jiang$^1$, Chen Chen$^1$, Ruihai Hou$^2$, Dong Wang$^{1*}$\thanks{
This work was supported by the National Natural Science Foundation of China under Grant No.62171250.}}
\address{
  $^1$Center for Speech and Language Technologies, BNRist, Tsinghua University, China  \\
  $^2$School of Artificial Intelligence, Beijing University of Posts and Telecommunications, China}
\email{$^*$Corresponding authors: wangdong99@mails.tsinghua.edu.cn}

\begin{document}

\maketitle

\begin{abstract}

Audio-visual person recognition (AVPR) has received extensive attention.
However, most datasets used for AVPR research so far are collected in constrained environments,
and thus cannot reflect the true performance of AVPR systems in real-world scenarios.
To meet the request for research on AVPR in unconstrained conditions,
this paper presents a multi-genre AVPR dataset collected `in the wild', named CN-Celeb-AV.
This dataset contains more than 419k video segments from 1,136 persons from public media.
In particular, we put more emphasis on two real-world complexities: (1) data in multiple genres; (2) segments with partial information.
A comprehensive study was conducted to compare CN-Celeb-AV with two popular public AVPR benchmark datasets,
and the results demonstrated that CN-Celeb-AV is more in line with real-world scenarios and
can be regarded as a new benchmark dataset for AVPR research. The dataset also involves a development set
that can be used to boost the performance of AVPR systems in real-life situations.
The dataset is free for researchers and can be downloaded from \emph{http://cnceleb.org/}.

\end{abstract}

\section{Introduction}
\label{sec:intro}

Biometric recognition is an automatic process of measuring and analyzing human biometrics and authenticating personal identity~\cite{bhattacharyya2009biometric}.
Voice and face are among the most popular biometrics, partly because they can be collected remotely and non-intrusively.
In the past few years, with the emergence of deep learning and data accumulation,
performance of the two biometric recognition techniques, i.e., speaker recognition (SR) and face recognition (FR), has been remarkably improved
and a wide range of applications has been fostered~\cite{variani2014deep,snyder2017deep,li2017deep,parkhi2015,deng2018arcface}.

Despite the impressive progress, either SR or FR suffers from their respective practical difficulties.
For audio-based SR, challenges are content variation, channel discrepancy, additive noise, spontaneous speaking styles, and even the change in physiological status.
For video-based FR, challenges can be from varied illumination, changed position, and accident occlusion.
Researchers have a long-term journey to combine the two modalities, primarily motivated by the
fact that audio and visual information are complementary, as has been demonstrated by some psychological experiments~\cite{sumby1954visual, partan1999communication,golumbic2013visual}.

For instance, with the assistance of facial information, humans could do significantly better on auditory perception tasks than if only audio signals were available.
Take another example, when a person wears a mask, the face is obscured, which increases both false acceptance and false rejection of face recognition; while this mask has a slight effect on the voice, making speaker recognition more suitable for person authentication.
A straightforward idea is to integrate the complementary information of audio and visual modalities to construct an audio-visual person recognition (AVPR) system,
that is supposed to be more robust, especially in \emph{unconstrained} conditions~\cite{chibelushi2002review}.
To answer such requirement for multi-modality person recognition, NIST kicked off an audio-visual challenge~\cite{sadjadi20202019} in SRE 2019,
and keep the effort in SRE 2021~\cite{sadjadi20222021}.

Existing AVPR research takes two approaches.
The hybrid approach trains SR and FR models separately and combines their decisions by either
score fusion~\cite{sell2018audio,alam2020analysis} or embedding integration~\cite{shon2019noise}.
An advantage of this approach is that the SR and FR models can be built using techniques and databases developed in their respective fields,
though the shortcoming is that the intrinsic relation between the two modalities is not fully utilized.
The second approach is multi-modal joint modeling.
For example, Qian et al.~\cite{qian2021audio} designed audio-visual neural nets based on either feature-level or embedding-level concatenation.
Tao et al.~\cite{tao2020audio} proposed a cross-modal discriminative neural net to learn shared audio-visual embeddings, and use
them to enhance a score-fusion AVPR.

All the above research reported promising performance.
For example, Qian et al.~\cite{qian2021audio} reported an EER reduction from 3.04 (V) and 1.62 (A) to 0.55 (A+V) on the VoxCeleb1 test set.
However, all these studies are based on relatively \emph{constrained} data, for which we mean (1) the data is clean and involves limited variations; (2) the information of the two modalities is fully exposed.
We argue that the results based on this type of data cannot reflect real-life complexity,
e.g., in  \emph{unconstrained} environments where information of one or two modalities is corrupted or lost.

To facilitate AVPR research in solving real-world challenges, we publish a new AVPR dataset named CN-Celeb-AV in this paper.
The new dataset follows the principles of CN-Celeb~\cite{li2022cn} in data collection but contains both audio and visual data. The entire dataset consists of two parts,
one is the `full-modality' part that involves full AV information and one is the `partial-modality' part that 
involves a large proportion of video segments whose audio or visual modality is corrupted or missing.
The entire dataset covers 11 different genres in real-world scenarios (as in CN-Celeb1/2) and contains more than 419k video segments from 1,136 persons (e.g., Chinese celebrities, video bloggers, and amateurs).
We hope that these distinct properties permit CN-Celeb-AV a suitable benchmark evaluation set for AVPR with real-world complexity.

\begin{table*}[htbp!]
\centering
\caption{Comparison of existing AVPR datasets. * means information is unavailable.}
\scalebox{0.94}{
  \begin{tabular}{llllllll}
    \toprule
    \textbf{Dataset}  & \textbf{\# of Spks} & \textbf{\# of Segments} & \textbf{\# of Hours}   & \textbf{Genre}       & \textbf{Uncertainty}    & \textbf{Status}  \\
    M2VTS~\cite{pigeon1997m2vts}       & 37          & *             & *                      & indoor               & constrained            & Private \\
    XM2VTS~\cite{messer1999xm2vtsdb}   & 295         & *             & *                      & indoor               & constrained            & Not Free  \\
    VidTIMIT~\cite{sanderson2002vidtimit} & 43       & 430           & 0.5   & indoor               & constrained            & Public \\
    \midrule
    MOBIO~\cite{mccool2012bi}           & 150        & 28,800        & 61    & indoor               & semi-constrained       & Public \\
    MSU-AVIS~\cite{chowdhury2018msu}    & 50         & 2,260         & 3     & indoor               & semi-constrained       & Public \\
    AveRobot~\cite{marras2019averobot}  & 111        & 2,664         & 5     & indoor               & semi-constrained       & Public \\
    WeCanTalk\cite{jones2022wecantalk} & 202        & 3,199         & *                      & telephone and video  & semi-constrained        & Private \\
    \midrule
    VoxCeleb1~\cite{nagrani2017voxceleb} & 1,251      & 153,516       & 352   & mostly interview     & unconstrained           & Public \\
    VoxCeleb2~\cite{chung2018voxceleb2} & 6,112      & 1,128,246     & 2,442 & mostly interview     & unconstrained           & Public \\
    JANUS (CORE)~\cite{sell2018audio}    & 360        & 1,593         & *                      & multi genres         & unconstrained           & public \\
    VAST (SID)~\cite{tracey2018vast}     & 300        & *             & *                      & multi genres         & unconstrained           & Private \\
    CN-Celeb-AV (\textbf{Ours})         & 1,136      & 419,663       & 669    & multi genres      & unconstrained         & Public  \\
    \bottomrule
  \end{tabular}
  \label{tab:comp}}
\end{table*}

\section{Review of AVPR datasets}
\label{lab:rev}

This section presents a brief review of existing AVPR datasets and compares them with CN-Celeb-AV.
The main information is summarized in Table~\ref{tab:comp}, and the following are some details.

Early work of AVPR focused on \emph{constrained} conditions where speech content, face pose, environment, and devices are strictly controlled.
The representative datasets in this stage involve clear faces and clean voices,
such as M2VTS~\cite{pigeon1997m2vts}, XM2VTS~\cite{messer1999xm2vtsdb}, and VidTIMIT~\cite{sanderson2002vidtimit}.
Further studies considered \emph{semi-constrained} conditions, in which
faces may be occluded and voices may be corrupted by noise or non-target speech,
but the recording environment and speech content are often controlled.
Typical datasets in this stage include MOBIO~\cite{mccool2012bi}, MSU-AVIS~\cite{chowdhury2018msu}, AveRobot~\cite{marras2019averobot} and WeCanTalk~\cite{jones2022wecantalk}.

Recently, the research focus has shifted to \emph{unconstrained} conditions.
A key feature of recognition tasks in these conditions is that the recording environment and device are fully unconstrained, and the target person may be unaware of being recorded.
VoxCeleb~\cite{nagrani2017voxceleb,chung2018voxceleb2} is a typical dataset with such features.
The data was collected from YouTube videos that were recorded by diverse devices in various conditions for different purposes.
A key issue of VoxCeleb is that the videos are mostly from interview programs which are a relatively simple genre
and not fully `in the wild'.

CN-Celeb-AV approaches real-world complexities in two aspects: it involves data from multiple genres, and audio and visual information may be partially available.
Some recent datasets involve multi-genre data, e.g.,
JANUS Multimedia dataset~\cite{sell2018audio} and VAST (SID)~\cite{tracey2018vast}, but they are either small or in private status.

\section{CN-Celeb-AV}

\subsection{Data description}

The purpose of the CN-Celeb-AV dataset is to evaluate the true performance of AVPR techniques in unconstrained conditions and provide a standard benchmark for AVPR research.
All the data was collected from Bilibili\footnote{https://www.bilibili.com/}, a popular Chinese public media. 
In total, it contains more than 419k video segments (669 hours) from 1,136 people (mostly Chinese celebrities) 
and covers 11 genres as in CN-Celeb~\cite{fan2020cn}.

Specifically, CN-Celeb-AV is composed of two parts.
The first `full-modality' part is the audio-visual version of the audio-only CN-Celeb1~\cite{fan2020cn} dataset.
Most of the data in this part contains both audio and visual information.
It is split into a development set and an evaluation set, which involve 689 people and 197 people respectively,
following the original split of CN-Celeb1~\cite{fan2020cn}. 
The two sets are denoted by CNC-AV-Dev-F and CNC-AV-Eval-F respectively, where `F' means `full modality'.

The second `partial-modality' part is a set of newly collected data that involves a large proportion of
video segments whose audio or visual information is corrupted or fully lost.
For example, the face and/or the voice of the target person may disappear shortly, be corrupted by noise, or even be fully unavailable. We denote this set of data by CNC-AV-Eval-P, indicating that it is an evaluation set, and with partial AV information.
It involves 308k video segments (427 hours) collected from 250 people. The duration of each video segment is 5 seconds, and the number of genres of each person is more than 3.

Table~\ref{tab:prof} presents the data profile of CN-Celeb-AV, and Table~\ref{tab:genre} presents the data distribution over genres. Note that the persons in CNC-AV-Eval-P were removed from CNC-AV-Dev-F and CNC-AV-Eval-F, which is the reason why the data in these two sets are smaller than those in the corresponding development and evaluation sets of CN-Celeb1.

\begin{table}[htbp]
    \centering
    \caption{The data profile of CN-Celeb-AV}
    \label{tab:prof}
    \scalebox{0.82}{
    \begin{tabular}{llll}
    \toprule
                          & CNC-AV-Dev-F   &  CNC-AV-Eval-F  & CNC-AV-Eval-P   \\
    \midrule
    \# of Genres          & 11             &  11             &  11             \\
    \# of Persons         & 689            &  197            &  250            \\
    \# of Segments        & 93,973         &  17,717         &  307,973        \\
    \# of Hours           & 199.70         &  41.96          &  427.74         \\
    \bottomrule
    \end{tabular}}
\end{table}

\begin{table}[htbp]
    \centering
    \caption{The distribution over genres of CN-Celeb-AV}
    \label{tab:genre}
    \scalebox{0.83}{
    \begin{tabular}{llll}
    \toprule
    Genres          & \# of Spks & \# of Segments  & \# of Hours \\
    \midrule
    Overall         & 1,136      & 419,663         & 669.36     \\
    \midrule
    Advertisement   & 45         & 3,888           & 5.44       \\
    Drama           & 136        & 7,118           & 7.70       \\
    Entertainment   & 555        & 40,660          & 59.98      \\
    Interview       & 907        & 155,457         & 261.96      \\
    Live Broadcast  & 269        & 57,979          & 84.61       \\
    Movie           & 66         & 2,405           & 3.01        \\
    Play            & 104        & 6,485           & 8.27       \\
    Recitation      & 45         & 3,457           & 5.99       \\
    Singing         & 384        & 24,752          & 43.98      \\
    Speech          & 214        & 61,503          & 109.25    \\
    Vlog            & 155        & 55,959          & 79.20      \\
    \bottomrule
    \end{tabular}}
\end{table}

\subsection{Features of CN-Celeb-AV}

As have been shown in Table~\ref{tab:comp}, CN-Celeb-AV possesses several desired features that make it suitable for
AVPR research to tackle real-world challenges.

\begin{itemize}
 \item Almost all the video segments involve real-world uncertainties, e.g., content, noise, channel, overlap, speaking style variations in voices
 and pose, illumination, expression, resolution, and occlusion in faces.
 \item There is a large proportion of single-speaker multi-genre data from many media files, allowing cross-genre and cross-session tests.
These tests match real-world situations.
 \item Information is partially observed in some video segments, especially in CNC-AV-Eval-P, making it suitable for testing the performance of AVPR systems in real-life complex conditions, which is the situation where AV multi-modal techniques are expected to get the most value.
\end{itemize}

\subsection{Collection pipeline}

The collection pipeline of CNC-AV-Dev-F and CNC-AV-Eval-F is the same as in CN-Celeb1~\cite{fan2020cn}. For CNC-AV-Eval-P, the situation is more complex as video segments with partial information need to be collected, leading to an increased burden on both automatic collection and human annotation. We therefore designed a simplified pipeline that made the data collection highly efficient.

\begin{figure}[htb!]
  \centering
  \includegraphics[width=0.75\linewidth]{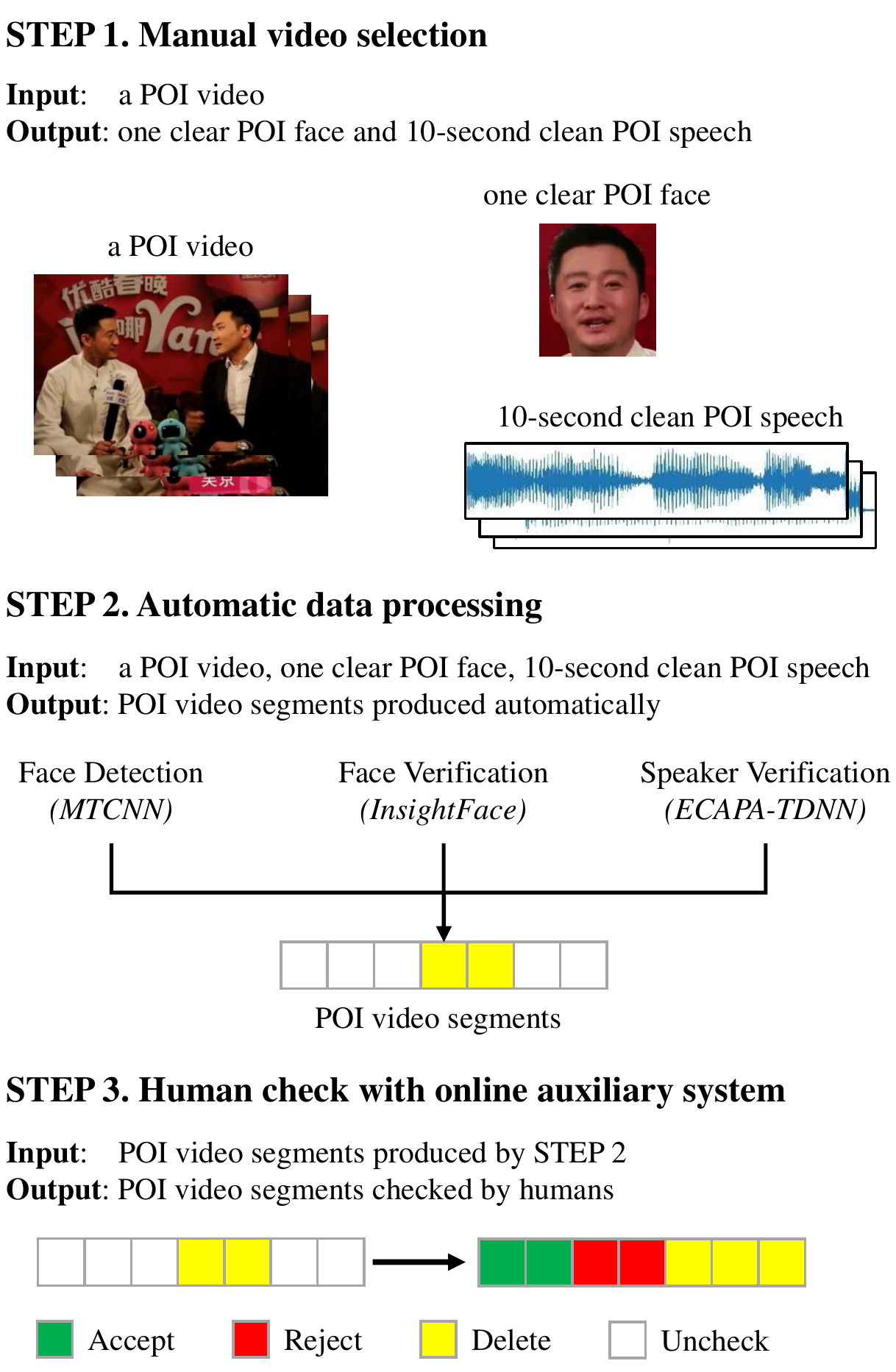}
  \caption{Illustration of the collection pipeline.}
  \label{fig:pipe}
\end{figure}

CNC-AV-Eval-P was collected in three steps: (1) candidate videos were manually selected; (2) an automatic process to extract
candidate segments; (3) human check to confirm valid segments.
This process is much faster than a purely human-based annotation and also avoids potential errors caused by a purely automated process.
We highlight that human check is important in our case: because the multi-genre data is very complex, the automatic process often makes mistakes.
We also developed a user-friendly crowd-sourcing platform to assist with video annotation and human check.
The source code of this platform will be published on the dataset webpage to help readers collect their own data.
The illustration of the collection pipeline is presented in Figure~\ref{fig:pipe} and the corresponding steps are summarized as follows.

\begin{itemize}
   \item \textbf{Step 1. Manually video selection.}
   (a) Select a person of interest (POI) who can be a celebrity or an uploader on Bilibili (a platform similar to YouTube),
   and then select multiple videos of the person as candidate videos. Annotate the genre type of each video.
   (b) For each video, grab a clear face image of the POI (POI face). Further select 10 short speech segments, each being 1-second long and containing the clear voice of the POI.
   Merge the 10 short segments into a 10-second POI speech. The POI face and POI speech are used in STEP 2 to select video segments containing the POI.
   Note that all the work in this step is performed by humans.
   \item \textbf{Step 2. Automatic data processing.}
   (a) Divide each candidate video into short segments of 5 seconds. Each segment contains 5 images and 500 speech frames.
   (b) For each segment, use the MTCNN model~\cite{zhang2016joint} to perform face detection and alignment,
   and then use the InsightFace model~\cite{deng2018arcface} to perform face verification. During the verification, the POI face and the face in each image are compared, and the maximum cosine distance is used as the detection score.
   (c) For each segment, use the ECAPA-TDNN model~\cite{desplanques2020ecapa} from the SpeechBrain toolkit~\cite{ravanelli2021speechbrain} to perform speaker verification,
   using the POI speech obtained in Step 1 as the enrollment speech.
   (d) Delete segments whose detection scores fall in the lowest 15\% with both the audio and visual modalities.
       The deleted segments are shown as yellow blocks in the annotation platform. All the rest segments are shown as white.
   \item \textbf{Step 3. Human check.}
   (a) Human annotators check each segment and label them as accepted (shown as green) or rejected (shown as red).
   The acceptance criterion is that humans can tell POI's existence with information from at least one modality.
   (b) To reduce human workload, an auxiliary system was designed to label low-confidence segments (shown as yellow) and update the
    threshold for yellow segments after each human annotation.
   (c) Double-check all the accepted segments by a senior annotator to guarantee data quality.
\end{itemize}

It is worth mentioning that in STEP 2, the audio and visual modalities are processed in parallel and symmetrically, thus avoiding bias towards a particular modality.
As far as we know, many existing datasets did not consider this modality bias.

\section{Experiments}

\subsection{Data}

In this section, we conduct AVPR experiments with two CN-Celeb-AV evaluation sets and other two popular datasets: MOBIO~\cite{mccool2012bi} and VoxCeleb1~\cite{nagrani2017voxceleb}.
We choose these two datasets for two reasons, one is that they have been widely used by researchers, and the other is that 
they released official trials, making the comparison easy.
For the two comparative datasets, we use the official trials defined by the data provider.
For CNC-AV-Eval-F/P, we choose video segments from a genre without too much complexity (such as interview and speech) for each POI.
The POI face and speech of this video are used as enrollment data for that POI, and all the remaining video segments are used for testing.

We argue that this trial design is in line with real-world scenarios where users
are often enrolled in a constrained condition while tested in unconstrained environments.
Due to the large amount of test utterances in CNC-AV-Eval-P, we construct the negative pairs for this evaluate set by a sampling approach.
Specifically, for each POI, we randomly sample negative pairs three times the positive pairs. 
The trials of the 4 datasets are listed in Table~\ref{tab:trial}.

\begin{table}[htb!]
\centering
\caption{The profile of 4 evaluation sets}
\label{tab:trial}
\scalebox{0.89}{
\begin{tabular}{llll}
\toprule
Dataset       & \# of Spks   & \# of Target   & \# of Nontarget  \\
\midrule
MOBIO         & 58           & 6,090      & 187,530    \\
VoxCeleb1-O   & 40           & 18,860     & 18,860     \\
CNC-AV-Eval-F & 197          & 17,693     & 53,079  \\
CNC-AV-Eval-P & 250          & 307,973    & 923,635 \\
\bottomrule
\end{tabular}}
\end{table}

\subsection{Model setting}

\subsubsection{Speaker verification}

We employ the ECAPA-TDNN model~\cite{desplanques2020ecapa} in the SpeechBrain toolkit~\cite{ravanelli2021speechbrain} for speaker verification.
The model is trained with VoxCeleb1.dev and VoxCeleb2.
All the speech data are first preprocessed by VAD and then fed into ECAPA-TDNN to extract speaker embeddings.
The cosine similarity is used to score the trials.

\subsubsection{Face verification}

To perform face verification, we sample images every 25 frames.
For each sampled image, RetinaFace~\cite{deng2020retinaface} is used to perform face detection and InsightFace~\cite{deng2018arcface}
is used to extract face embeddings, and verification scores are based on cosine similarity.
Since there are multiple face images in both enrollment and test videos,
a pooling scheme is required to make full use of these images, either at the embedding level or
the score level.

For MOBIO and VoxCeleb1.O, only the target person may appear in the video.
We therefore simply average the face embeddings of all the sampled images to represent a video
for either enrollment or test.



For CNC-AV-Eval-F/P, the POI face is used for enrollment.
During the test, multiple faces may appear in the test video. 
We compare the person vector to the face 
embedding of every sampled image, and the maximum cosine similarity 
is used as the verification score.

\subsubsection{System fusion}

Simple score fusion is used to fuse the audio and visual modalities. Calibration is
a standard technique to perform such fusion. In a nutshell, calibration maps
raw scores produced by a decision system to log-likelihood ratios (LLRs)~\cite{brummer2006application},
The LLRs have a clear probabilistic interpretation, making them theoretically suitable for combining decisions from
different systems~\cite{kittler2007quality}. A CLLR-based calibration routine implemented in the BOSARIS toolkit~\cite{brummer2013bosaris}
is used to perform calibration. Once the scores of the visual and audio streams are calibrated, we simply average them to get the final score.

\subsection{Results}

The results in terms of EER(\%) and minDCF($P_{tar}$ = 0.1) are reported in Table~\ref{tab:res}.
Firstly, we can see that both single-modal and multi-modal systems achieve good performance
on the MOBIO and VoxCeleb1 datasets.
This is expected since the information is almost complete in these two datasets and
the interference is limited.
In contrast, the performance on the two CNC-AV-Eval datasets is much worse, especially
with the visual modality. This is also not surprising, as the
data in these datasets is more complex, e.g., multiple faces may occur in the same 
frame and the target face may be small or occluded. Since all these complexities occur 
in real-life conditions, the results indicate that the present person recognition techniques 
are still far from perfect, either with audio or visual clues. 

Secondly, it can be observed that the performance of multi-modal systems is consistently
better than single-modal systems on all the datasets, demonstrating the benefit of
multi-modal processing. However, even with the AV models, the performance of the two CNC-AV-Eval
datasets are still poor, suggesting further research.

\begin{table}[htb!]
\centering
\caption{Results on different evaluation sets.}
\label{tab:res}
\scalebox{0.89}{
\begin{tabular}{l|cccc}
\toprule
&\multicolumn{3}{c}{EER(\%) / minDCF($P_{tar}$ = 0.1)}\\
\midrule
   & Audio            & Visual            & Fusion  \\
\midrule
MOBIO          & 2.48/0.146     & 0.79/0.035      & 0.25/0.009   \\
VoxCeleb1-O    & 1.04/0.057     & 1.89/0.088      & 0.30/0.011    \\
CNC-AV-Eval-F  & 14.98/0.458    & 20.30/0.597     & 11.94/0.371   \\
CNC-AV-Eval-P  & 16.71/0.449    & 17.19/0.460     &  9.04/0.287   \\
\bottomrule
\end{tabular}}
\end{table}

In the last experiment, we test the value of the released development set CNC-AV-Dev-F.
We simply use this set to train two LDA models that project the audio and visual 
embeddings respectively.  For the audio stream, it reduces the dimension of speaker embeddings 
from 192 to 128, and for the visual stream, it reduces the dimension of face embeddings 
from 512 to 128. The projected embeddings are then used to perform tests as in the previous experiment.
The results are shown in Table~\ref{tab:res2}. It can be seen that significant performance gains
were obtained, with both the two single-model systems and the fusion system. This demonstrated the
value of CNC-AV-Dev-F. 

\begin{table}[htb!]
\centering
\caption{Results on the two CNC-AV evaluation datasets with LDA trained on the CNC-AV development set.}
\label{tab:res2}
\scalebox{0.89}{
\begin{tabular}{l|cccc}
\toprule
&\multicolumn{3}{c}{EER(\%) / minDCF($P_{tar}$ = 0.1)}  \\
\midrule
  & Audio        & Visual          & Fusion  \\
\midrule
CNC-AV-Eval-F  & 10.63/0.358    & 18.97/0.487     &  8.64/0.271   \\
CNC-AV-Eval-P  & 13.90/0.387    & 15.51/0.431     &  7.40/0.243   \\
\bottomrule
\end{tabular}}
\end{table}

\section{Conclusion}

We introduced CN-Celeb-AV, a free multi-modal dataset for audio-visual person recognition research.
This dataset consists of one development set and two evaluation sets. The two evaluation sets 
were designed to represent test conditions with
full-modality and partial-modality information respectively.
We compared the two evaluation sets with two existing AV datasets, MOBIO and VoxCeleb1. Experimental results demonstrated that CN-Celeb-AV represents more challenging real-life situations, and 
simply employing and combining the current SOTA speaker and face recognition models 
cannot achieve satisfactory results in this condition.
Finally, we verified that the development set of CN-Celeb-AV can be used to improve the performance of 
AVPR system in real-life conditions.


\newpage

\bibliographystyle{IEEEtran}
\bibliography{mybib}

\end{document}